\documentclass[runningheads]{llncs}
\usepackage{cite}
\usepackage{amsmath,amssymb,amsfonts}
\usepackage{algorithmic}
\usepackage{graphicx}
\usepackage{textcomp}
\usepackage{xcolor}
\usepackage{standalone}

\usepackage{scalerel}
\usepackage{tikz}
\usetikzlibrary{calc, shapes, arrows, positioning, patterns, decorations.pathreplacing, backgrounds, petri, automata, shadows, external, trees}
\usepackage{pgf}
\usepackage{pgfplots}
\usepackage{svg}

\usepackage{cleveref}

\usepackage{url}

\usepackage{caption}
\usepackage{subcaption}

\usepackage{gincltex}

\newcommand{\actuniverse}[0]{\mathcal{A}} 
\newcommand{\lpmuniverse}[0]{\mathbb{U}_{\lpm}} 
\newcommand{\clustersetuniverse}[0]{\mathbb{U}_{\partition}}

\newcommand{\set}[1]{\{ #1 \}} 
\newcommand{\multiset}[1]{[\hspace{0.1em} #1 \hspace{0.1em}]} 
\newcommand{\seq}[1]{\langle #1 \rangle} 

\newcommand{\powerset}[1]{\mathcal{P}(#1)}
\newcommand{\allmultiset}[1]{\mathbb{M}(#1)} 
\newcommand{\allsequence}[1]{#1^*}
\newcommand{\proj}[2]{#1_{\upharpoonright_{#2}}} 
\newcommand{\partmapsto}[0]{\not\to} 
\newcommand{\st}[0]{\vert} 
\newcommand{\length}[1]{\vert #1 \vert} 
\newcommand{\seqvar}[0]{\sigma}

\newcommand{\emarking}[0]{\multiset{}} 
\newcommand{\imarking}[0]{M_i}
\newcommand{\fmarking}[0]{M_f}

\newcommand{\lang}[1]{\mathcal{L}(#1)} 
\newcommand{\nlang}[2]{\mathcal{L}^{#2}(#1)}
\newcommand{\preset}[1]{\bullet #1} 
\newcommand{\postset}[1]{#1 \bullet} 

\newcommand{\fire}[1]{\xrightarrow{#1}} 
\newcommand{\fireseq}[0]{\sigma} 
\newcommand{\allfireseq}[2][]{\mathcal{F}_{#1}(#2)} 

\newcommand{\trace}[0]{\rho}
\newcommand{\eventlog}[0]{L}

\newcommand{\lpm}[0]{\textit{LPM}}
\newcommand{\lpmvar}[0]{\textit{lpm}}
\newcommand{\allfireseqlpm}[1]{\allfireseq[\lpm]{#1}}
\newcommand{\modelA}[0]{\lpmvar_A}
\newcommand{\modelB}[0]{\lpmvar_B}
\newcommand{\rank}[1]{\textit{rank}_{#1}}
\newcommand{\freetransitions}[0]{T_{in}}
\newcommand{\rankingfuncuniverse}[0]{\mathbb{U}_{\textit{rank}}}

\newcommand{\simm}[1]{\mathit{sim}_{#1}}
\newcommand{\simmeasurevar}[0]{sim} 
\newcommand{\simmeasureuniverse}[0]{\mathbb{U}_{sim}} 
\newcommand{\partition}[0]{\sqcap}

\newcommand{\constname}[1]{\textit{#1}}
\newcommand{\lpmtext}[0]{LPM}

\begin{document}
\title{Grouping Local Process Models}
%
%
\author{Viki Peeva \and
Wil M.P. van der Aalst}
\authorrunning{Viki Peeva and Wil M.P. van der Aalst}
%
\institute{Chair of Process and Data Science (PADS) \\ RWTH Aachen University,
Aachen, Germany \\ \email{\{peeva, wvdaalst\}@pads.rwth-aachen.de}}
\maketitle              
\begin{abstract}
In recent years, process mining emerged as a proven technology to analyze and improve operational processes. An expanding range of organizations using process mining in their daily operation brings a broader spectrum of processes to be analyzed. 
Some of these processes are highly unstructured, making it difficult for traditional process discovery approaches to discover a start-to-end model describing the entire process. Therefore, the subdiscipline of Local Process Model (\lpmtext{}) discovery tries to build a set of LPMs, i.e., smaller models that explain sub-behaviors of the process. However, like other pattern mining approaches, LPM discovery algorithms also face the problems of model explosion and model repetition, i.e., the algorithms may create hundreds if not thousands of models, and subsets of them are close in structure or behavior. This work proposes a three-step pipeline for grouping similar LPMs using various process model similarity measures.
We demonstrate the usefulness of grouping through a real-life case study, and analyze the impact of different measures, the gravity of repetition in the discovered LPMs, and how it improves after grouping on multiple real event logs.

\end{abstract}

\begin{sloppypar}
\section{Introduction}\label{sec:introduction}

\begin{figure}[t]
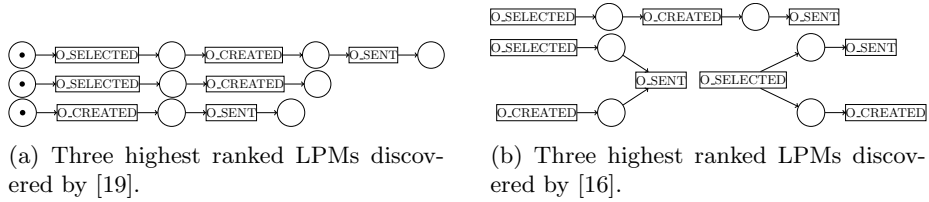

    \centering
    \begin{subfigure}[b]{0.475\textwidth}
    \includegraphics[width=\linewidth]{fig_bpi2012_res10939_tax_lpm1.tex}
    \\
    \includegraphics[width=0.74\linewidth]{fig_bpi2012_res10939_tax_lpm2.tex}
    \\
    \includegraphics[width=0.68\linewidth]{fig_bpi2012_res10939_tax_lpm3.tex}
    \caption{Three highest ranked \lpmtext{}s discovered by \cite{DBLP:journals/jides/TaxSHA16}.}
    \end{subfigure}
    \hfill
    \begin{subfigure}[b]{0.475\textwidth}
    \includegraphics[width=0.8\linewidth]{fig_bpi2012_res10939_peeva_lpm1.tex}
    \\
    \includegraphics[width=0.45\linewidth]{fig_bpi2012_res10939_peeva_lpm2.tex}
    \hfill
    \includegraphics[width=0.52\linewidth]{fig_bpi2012_res10939_peeva_lpm3.tex}
    \caption{Three highest ranked \lpmtext{}s discovered by \cite{DBLP:conf/apn/PeevaMA22}.}
    \label{fig:similar_models_peeva}
    \end{subfigure}
    \caption{\lpmtext{}s discovered using \cite{DBLP:journals/jides/TaxSHA16} and \cite{DBLP:conf/apn/PeevaMA22} on the filtered and transformed \textit{BPIC2012} event log for resource $10939$ as explained in \cite{DBLP:journals/jides/TaxSHA16}.}
    \label{fig:similar_models}
\end{figure}

Process mining is a scientific discipline for discovering, monitoring, and improving processes via readily-available data from different data management systems. The three main pillars of process mining are process discovery, conformance checking, and process enhancement \cite{DBLP:books/sp/Aalst16}. 
As the interest in process mining grows, new applications pose new challenges.
One such challenge is discovering a single start-to-end model for highly unstructured processes (\cite[Figure~7]{DBLP:journals/jides/TaxSHA16}). To resolve this problem, one usually focuses on frequent behavior because of the $80/20$ rule of data variability. However, in some domains, especially ones covering human behavior, the rule does not hold and better solutions are needed. A relatively new field is Local Process Model (\lpmtext{}) discovery \cite{DBLP:journals/jides/TaxSHA16}, where the idea is to build smaller process models explaining fragments of the behavior instead of one overall model. Yet, current \lpmtext{} discovery approaches return hundreds or thousands of models for one event log (\emph{model explosion}), with highly similar models repeating between them (\emph{model repetition}) as shown in \Cref{fig:similar_models}. Although this is desirable in some use cases \cite{DBLP:conf/sac/DelcoucqLFA20,DBLP:conf/emisa/MannhardtT17}, it does not help with a better understanding of highly variable processes.

To alleviate this problem, in this work, we propose a pipeline that groups similar \lpmtext{}s, and for each group, one representative \lpmtext{} is chosen. By doing so, highly similar models that describe small differences in behavior are grouped together, allowing analysts to focus on the bigger picture by examining a smaller diverse set of \lpmtext{}s and only dive deeper into the differences when necessary. More specifically, we start with an \lpmtext{} discovery approach that returns a set of \lpmtext{}s. These \lpmtext{}s are then clustered such that model similarity, using different process model similarity measures, is measured, and for each cluster, a representative \lpmtext{} is chosen. We evaluate the proposed approach on multiple event logs by comparing model diversity between the originally returned set of \lpmtext{}s and the representative set obtained after grouping. Additionally, we demonstrate the benefit of clustering \lpmtext{}s by inspecting a smaller set of models on the \constname{BPIC2012-res10939} event log.

The rest of the paper is structured as follows. In \Cref{sec:preliminaries}, we introduce the needed preliminaries to follow the rest of the paper. \Cref{sec:local_process_models} introduces what \lpmtext{}s are and in \Cref{sec:pm_similarity_measures} we present different process model similarity measures. We explain the framework in \Cref{sec:evaluation_setup} and we present the obtained results in \Cref{sec:evaluation_results}. We conclude the paper with final remarks in \Cref{sec:conclusion}.

\section{Preliminaries}\label{sec:preliminaries}

\subsection{General}
We define sets ($X = \set{a,b}$), multisets ($ M = [a^2,b^3]$), sequences ($\seqvar = \seq{a,b,c}$), and tuples ($t = ( a,b,c )$) as usual. Given a set $X$, $\powerset{X}$ is the power set of $X$, $\allsequence{X}$ represents the set of all sequences over $X$, and $\allmultiset{X}$ is the set of all multisets over $X$. We use $\seqvar(i)$ to denote the $i$-th element of the sequence $\seqvar$ and $M(a) = 2$ to denote that item $a$ appears twice in the multiset $M$.

We use $f(X)=\{f(x) \vert x \in X\}$ (and $f(\seqvar)=\langle f(\seqvar(1)), f(\seqvar(2)),..., f(\seqvar(n)) \rangle$) to apply the function $f$ to every element in the set $X$ (the sequence $\seqvar$) and $\proj{f}{X}$ (respectively $\proj{\seqvar}{X}$) to denote the projection of the function $f$ (respectively the sequence $\seqvar$) on the set $X$.

\subsection{Process Mining}
Normally, the collected data used for process analysis is transformed in the form of \emph{event logs}. Hence, in \Cref{def:trace-log}, we formally define \emph{traces} and \emph{event logs}. Note that although traces are usually defined as sequences of events, in this work, we are only interested in the activity executed by each event.

\begin{definition}[Trace, Event Log]
    \label{def:trace-log}
    Given the universe of activities $\actuniverse$, we define $\trace \in \actuniverse^*$ as a \emph{trace}, and $\eventlog \in \allmultiset{\actuniverse^*}$ as an \emph{event log}.
\end{definition}

In \Cref{def:petri-net}, we define \emph{labeled Petri nets}. Note that a transition $t \in T$ with $l(t) = \tau$ is called silent and that there may be duplicate transitions $t_1, t_2 \in T$ such that $l(t_1) = l(t_2)$.

\begin{definition}[Labeled Petri net]
    \label{def:petri-net}
    A \emph{labeled Petri net} $N = (P, T, F, l)$ is a tuple, where $P$ is a set of places and $T$ is a set of transitions such that $P \cap T = \emptyset$. $F \subseteq (P \times T) \cup (T \times P)$ is the flow relation, and $l: T \to \actuniverse \cup \set{\tau}$ the labeling function.
\end{definition}

Now, given a node $x \in P \cup T$, we define the \emph{preset} of $x$ as $\preset{x} = \set{y \in P \cup T \st (y, x) \in F}$ and the \emph{postset} of $x$ as $\postset{x} = \set{y \in P \cup T \st (x, y) \in F}$. 

To attach behavior to labeled Petri nets we use a \emph{marking} and the \emph{firing rule}. A marking $M$ denotes the state of a Petri net as a multiset of places ($M \in \allmultiset{P}$) and the firing rule allows for changes between states (i.e., markings). Given a marking $M$, a transition $t$ is \emph{enabled} in the marking $M$ if and only if $\preset{t} \subseteq M$. If a transition $t$ is enabled in the marking $M$, it can fire and change the state of the net to a new marking $M' = (M \setminus \preset{t}) \cup \postset{t}$. We write $M \fire{t} M'$. A sequence of transitions $\fireseq = \seq{t_1, \dots, t_n} \in \allsequence{T}$ is enabled in $M$ and by firing it marking $M'$ is reached if and only if there exist $M_0$, $M_1$, $\dots$, $M_n$ such that $M_0=M$, $M_n=M'$, and $M_{i-1}\fire{t_i}M_i$ for $1 \leq i \leq n$. We write $M \fire{\fireseq} M'$.

Now, we formally define an \emph{accepting Petri net} in \Cref{def:accepting-petri-net} and the set of all its firing sequences in \Cref{def:accepting-petri-net-traces}.

\begin{definition}[Accepting Labeled Petri Net]
    \label{def:accepting-petri-net}
    An \emph{accepting labeled Petri net} is a triple $(N, \imarking, \fmarking)$ such that $N = (P,T,F,l)$ is a labeled Petri net, $\imarking \in \allmultiset{P}$ is the initial marking, and $\fmarking \in \allmultiset{P}$ is the final marking.
\end{definition}

\begin{definition}[Complete Firing Sequences]
    \label{def:accepting-petri-net-traces}
    Let $\textrm{AN} = (N, \imarking, \fmarking)$ be an accepting Petri net, $\allfireseq{\textrm{AN}} = \set{\fireseq \in \allsequence{T} \st \imarking \fire{\fireseq} \fmarking}$ is the set of complete firing sequences for $\textrm{AN}$.
\end{definition}

\section{Local Process Models}\label{sec:local_process_models}

In contrast to process discovery, whose task is to discover one model that explains the traces of an event log from start to end, \lpmtext{} discovery tries to mine a set of models each matching some particular sub-behavior represented by subsequences in the event log.

\lpmtext{}s were first introduced in \cite{DBLP:journals/jides/TaxSHA16} as a replacement for process discovery of highly unstructured processes. However, afterward, the use cases where \lpmtext{}s are used expanded (e.g., \cite{DBLP:conf/bpm/DeevaW18,DBLP:conf/sac/DelcoucqLFA20,DBLP:conf/ewgdss/KirchnerM18,DBLP:conf/otm/LeemansTH18,DBLP:journals/is/MannhardtLRAT18,DBLP:conf/emisa/MannhardtT17,PijnenborgVFLG21}) 
and multiple approaches and extensions for \lpmtext{} discovery followed \cite{DBLP:conf/caise/AcheliGW19,DBLP:conf/apn/PeevaMA22,DBLP:conf/icpm/BruningsF022}.
 
All existing approaches \cite{DBLP:conf/caise/AcheliGW19,DBLP:conf/apn/PeevaMA22,DBLP:journals/jides/TaxSHA16}, with the exception of \cite{DBLP:conf/icpm/BruningsF022}, suffer from the \emph{model explosion} problem. The approach in \cite{DBLP:conf/icpm/BruningsF022} first finds common subsequences and then discovers models on them. This, however, makes it predisposed to similar problems as the traditional process discovery approaches and restricts the set of \lpmtext{}s use-case scenarios the approach can be applied to. Still, as everyone else, they are prone to the \emph{model repetition} problem.
The approach in \cite{DBLP:journals/jides/TaxSHA16} incrementally extends process trees by adding new nodes, in \cite{DBLP:conf/caise/AcheliGW19} two existing \lpmtext{}s (represented with process trees) differing only in one node are joined together to create a larger \lpmtext{}. In the new model, the differing nodes are added as children to one of the process tree operators. Finally, in \cite{DBLP:conf/apn/PeevaMA22}, \lpmtext{}s are created by joining place nets together. This makes it clear that all approaches would build highly similar, i.e., repetitive models. Additionally, if one focuses on frequent behavior, the highly similar models would all explain highly similar behavior making highly-ranked models contain clusters of repetitive models.
The implementations of \cite{DBLP:journals/jides/TaxSHA16} and \cite{DBLP:conf/apn/PeevaMA22} offer a rudimentary grouping of the models. However, it becomes evident this is not sufficient when one considers human analysts manually inspecting and analyzing hundred of \lpmtext{}s. Therefore, with this work, we try to alleviate this shortcoming of \lpmtext{} discovery approaches.

In this work, we focus on \lpmtext{}s discovered with \cite{DBLP:conf/apn/PeevaMA22}. Therefore, although \lpmtext{}s can, in general, be represented by any modeling language (Petri nets, process trees, BPMNs, etc.), in this work we restrict to a subclass of accepting labeled Petri nets. We show a few example \lpmtext{}s in \Cref{fig:similar_models_peeva} and we give a formal definition in \Cref{def:lpm}. We use $\freetransitions = \set{t \in T \st \preset{t} = \emptyset}$ to denote the transitions that have an empty preset and we call them \emph{unrestricted transitions}.
We define the set of complete valid firing sequences of such \lpmtext{}s in \Cref{def:lpm-behavior} by restricting that each place in the net can receive at most one token from unrestricted transitions.

\begin{definition}[Local Process Models]
    \label{def:lpm}
    A \emph{Local Process Model} (\lpmtext{}) is an accepting labeled Petri net $\lpmvar = (N_{\lpmvar}, \imarking, \fmarking)$ such that $N_{\lpmvar}=(P, T, F, l)$ is a labeled Petri net that satisfies the following restrictions:
    \begin{enumerate}
        \item $\forall_{x, x' \in P \cup T} \, \exists_{\seq{x_1, \dots, x_n}} (x = x_1 \land x' = x_n \land \forall_{1\leq i < n} ((x_i, x_{i+1}) \in F) \lor (x_{i+1}, x_{i}) \in F))$, i.e., there is only one connected component, and
        \item $\forall_{p \in P}(\preset{p} \neq \emptyset \land \postset{p} \neq \emptyset)$, i.e., each place has at least one incoming and one outgoing arc,
    \end{enumerate}
    and $\imarking \in \allmultiset{P}$ and $\fmarking \in \allmultiset{P}$ are the initial and final marking. We use $\lpmuniverse$ to denote the universe of such \lpmtext{}s.
\end{definition}

\begin{definition}[Local Process Model Behavior]
    \label{def:lpm-behavior}
    Given an \lpmtext{} $\lpmvar=(N_{\lpmvar}, \imarking, \fmarking)$ such that $N_{\lpmvar}=(P, T, F,l)$, we define $\allfireseqlpm{\lpmvar} = \set{\fireseq \in \allfireseq{\lpmvar} \st \forall_{1 \leq i < j \leq \length{\fireseq}} (\fireseq(i) \in \freetransitions \land \fireseq(j) \in \freetransitions \implies \postset{\fireseq(i)} \cap \postset{\fireseq(j)} = \emptyset) }$
    to be all valid complete firing sequences of $\lpmvar$.
\end{definition}

The \emph{language} of an \lpmtext{} $\lpmvar$ is obtained by projecting all valid complete firing sequences on the transition labels and removing $\tau$-skips, i.e., $\lang{\lpmvar} = \set{\proj{l(\fireseq)}{\actuniverse} \st \fireseq \in \allfireseqlpm{\lpmvar}}$. We use $\nlang{\lpmvar}{n} = \set{\proj{l(\fireseq)}{\actuniverse} \st \fireseq \in \allfireseqlpm{\lpmvar} \land \length{\fireseq} \leq n}$ to denote the language restricting to complete firing sequences of length at most $n$. We can use the language to measure conformance with respect to an event log $L$ and rank the \lpmtext{}s. The ranking can take into consideration different quality measures, such as fitness, precision, and simplicity. We write $\rank{L} \in \lpmuniverse \partmapsto \mathbb{N}$ to denote a ranking function, and $\rankingfuncuniverse$ to denote the universe of all such ranking functions.

We later use these definitions to extract features from the \lpmtext{}s and formalize the different similarity measures.
\section{Process Model Similarity Measures}\label{sec:pm_similarity_measures}

To get an overview of existing similarity measures, we considered multiple survey papers \cite{DBLP:books/daglib/p/DijkmanD13,DBLP:journals/debu/DumasGD09,DBLP:journals/csur/SchoknechtTFOL17,DBLP:conf/bpm/ThalerSFO016,DBLP:journals/cii/BeckerL12}. Although there can be small differences in how they categorize different similarity measures, all of them agree,  the basic split is into measures that compare the structure of the process model and those that compare the behavior. Subsequently, one can consider the level of abstraction used, e.g., complete language versus weak order relations. Therefore, we choose five representative similarity measures. Before introducing the specific measures, we first define what a similarity measure is in \Cref{def:sim_measure}.

\begin{definition}[Similarity Measure]
    \label{def:sim_measure}
    A \emph{similarity measure} $\simm{\mathit{name}} \in \lpmuniverse \times \lpmuniverse \to [0,1]$ is a function that calculates the similarity between two \lpmtext{}s. We use $`\mathit{name}`$ to distinguish a specific measure, and $\simmeasureuniverse$ to denote the universe of all similarity measures.
\end{definition}

To introduce the similarity measures, we assume we are given two \lpmtext{}s $\modelA = (N^{A}_{\lpmvar}, \emarking, \emarking)$ s.t. $N^{A}_{\lpmvar}=(P_A, T_A, F_A, l_A)$ and $\modelB = (N^{B}_{\lpmvar}, \emarking, \emarking)$ s.t. $N^{B}_{\lpmvar}=(P_B, T_B, F_B, l_B)$. In the following, we illustrate the measures we use in this work with the help of $\modelA$ and $\modelB$.

\emph{Transition label comparison} is the most simple measure we investigate. The measure calculates the transition label overlap between the models.
\begin{equation*}
\simm{\textit{transition}}(\modelA, \modelB) = \frac{2 * \length{l_A(T_A) \cap l_B(T_B)}}{\length{l_A(T_A)}+\length{l_B(T_B)}}  
\end{equation*}

\emph{Node comparison} is somewhat more complex, in that it includes place overlap as well. We use this measure to represent structural measures using abstraction. The measure calculates the similarity between two models by combining transition label comparison and place matching between the nets. We assign to each pair of places a matching gain $g(p_1, p_2) = \frac{1}{2}*\frac{2 * \length{l_A(\preset{p_1}) \cap l_B(\preset{p_2})}}{\length{l_A(\preset{p_1})} + \length{l_B(\preset{p_2})}} + \frac{1}{2}*\frac{2 * \length{l_A(\postset{p_1}) \cap l_B(\postset{p_2})}}{\length{l_A(\postset{p_1})} + \length{l_B(\postset{p_2})}}$ and we use the Hungarian algorithm \cite{DBLP:books/daglib/p/Kuhn10} to solve the assignment problem. We use $G_{\textit{places}}$ to represent the gain of the optimal assignment. Then, we define the measure as 
\begin{equation*}
\simm{\textit{node}}(\modelA, \modelB)=\frac{2 * \length{l_A(T_A) \cap l_B(T_B)} + 2 * G_{\textit{places}}}{\length{l_A(T_A)}+\length{l_B(T_B)} + \length{P_A} + \length{P_B}}    
\end{equation*}

\emph{Eventually-follow graph similarity} is a behavioral abstraction measure that measures the overlap of the eventually-follows relation in the languages of the two models. We calculate it as 
\begin{equation*}
\simm{\textit{efg}}^n(\modelA, \modelB) = \frac{2 * |\textit{EF}^{\, n}_A \cap \textit{EF}^{\, n}_B|}{|\textit{EF}^{\, n}_A|+|\textit{EF}^{\, n}_B|}    
\end{equation*}
such that $\textit{EF}_A^{\, n} = \set{(a, b) \st \exists_{\trace \in \nlang{\modelA}{n}}(\exists_{1 \leq i < j \leq \length{\trace}}(a = \rho_i \land b = \rho_j))}$ and $\textit{EF}^{\, n}_B$ is defined correspondingly.

\emph{Full trace matching comparison} represents the more sophisticated behavioral measures. We define it as 
\begin{equation*}
    sim^n_{\textit{full}}(\modelA, \modelB) = \frac{2 * G_{\textit{traces}}}{\length{\nlang{\modelA}{n}} + \length{\nlang{\modelB}{n}}}
\end{equation*}
where $G_{\textit{traces}}$ represents the gain of the optimal trace assignment. To calculate the gain between two traces we invert the normalized Levenshtein distance.

Finally \emph{graph edit model comparison}
represents sophisticated structural measures. It calculates model similarity by using the graph edit distance (ged) as defined in \cite{DBLP:conf/icpram/Abu-AishehRRM15}, where the node substitution cost is $1$ if the nodes differ in type, i.e., one is a place and the other transitions, or if the compared nodes are differently labeled transitions. The node substitution cost between two places is calculated as $1 - g(p_1, p_2)$, where $g(p_1, p_2)$ is the gain defined as before. The edge substitution cost takes the average of the node substitution cost between the source and sink nodes of the two edges. To convert the ged to a similarity measure, we use the formula below. 
\begin{equation*}
    sim_{\textit{ged}}(\modelA, \modelB) = 1 - ged(\modelA, \modelB)
\end{equation*}

In the remainder, we also use the term distance measure, which we always consider to be the inverse of the similarity, i.e., $\textit{dist}_{\textit{name}}(\modelA, \modelB) = 1 - \simm{\textit{name}}(\modelA, \modelB)$ for any $\textit{name} \in \set{\textit{transition}, \textit{node}, \textit{efg}, \textit{full}, \textit{ged}}$

\section{Method to Group \lpmtext{}s}\label{sec:evaluation_setup}

\begin{figure}[t]
    \centering
    \includegraphics[width=\linewidth]{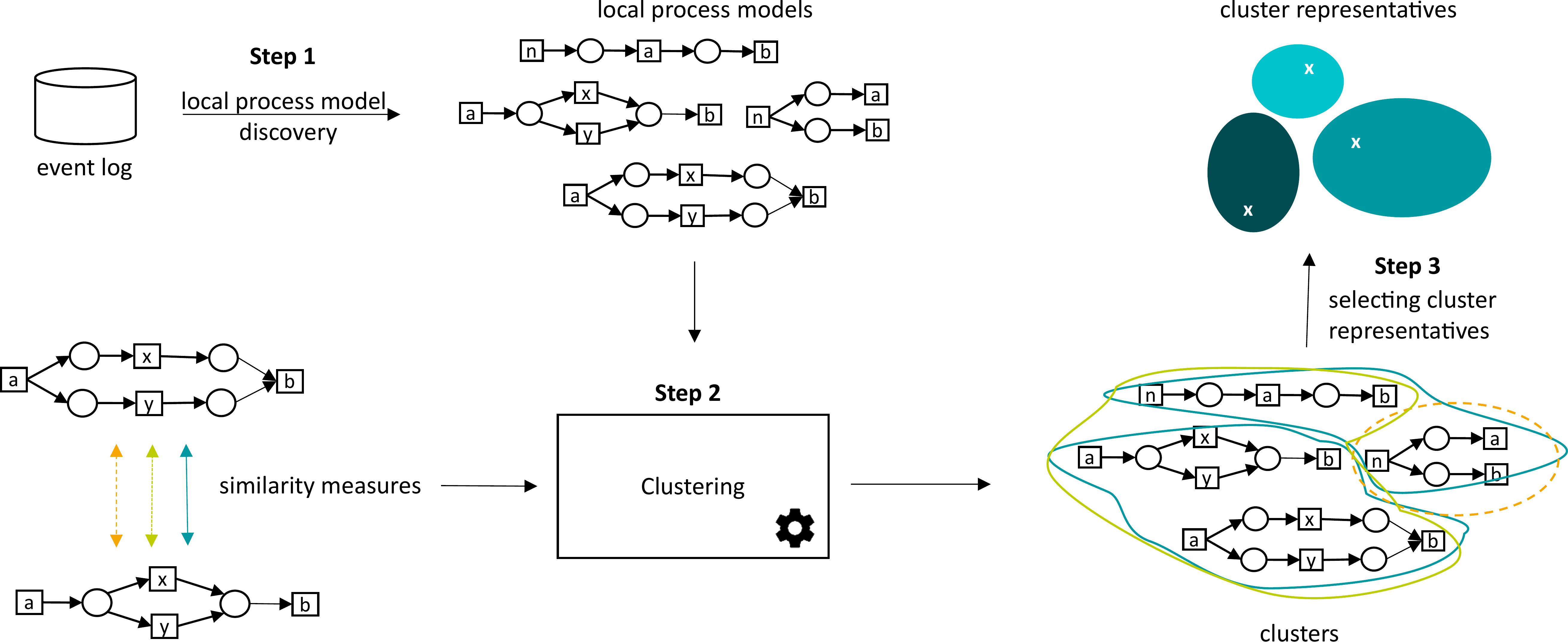}
    \caption{Illustration of the proposed three-step pipeline.}
    \label{fig:method}
\end{figure}

In this work, we propose a three-step pipeline that starts with an event log and a multitude of process model comparison measures and ends with groups of similar \lpmtext{}s, as shown in \Cref{fig:method}. The first step is discovering \lpmtext{}s (Step $1$), which can also be omitted, starting the pipeline with a set of \lpmtext{}s instead. Then, the models are clustered such that the similarity between them is determined by the previously defined process model similarity measures (Step $2$). Finally, for each cluster, we choose a representative model (Step $3$).

\subsection{Local Process Model Discovery (Step 1)}
In the first step, we focus on discovering a set of \lpmtext{}s $\lpm_L$ given an event log $L$. 
Although multiple approaches are available, in this work, we use the approach presented in \cite{DBLP:conf/apn/PeevaMA22}. 
The produced models are ranked from highest to lowest using a rank function $\textit{rank}_L$ as previously defined.

\subsection{Clustering (Step 2)}
In the clustering step, we accept a set of \lpmtext{}s and a similarity measure and return a set of clusters. We define the \emph{universe of \lpmtext{} cluster sets} in \Cref{def:cluster-set}, and a \emph{clustering algorithm} in \Cref{def:clustering}.

\begin{definition}[Universe of Local Process Model Cluster Sets]
    \label{def:cluster-set}
    We define $\clustersetuniverse = \set{X_{\lpm} \subseteq \powerset{\lpm_L} \st \lpm_L \subseteq \lpmuniverse \land \emptyset \not\in X_{\lpm} \land \bigcup X_{\lpm} = \lpm_L}$ to be the universe of \lpmtext{} cluster sets.
\end{definition}
\begin{definition}[Clustering Algorithm]
    \label{def:clustering}
     We define $\textit{clust} \in \powerset{\lpmuniverse} \times \simmeasureuniverse \partmapsto \clustersetuniverse$ to be a \emph{clustering algorithm}. To denote a clustering algorithm given some set of parameters $P$, we write $\textit{clust}_P$.
\end{definition}

The goal of the clustering algorithm is to return the \lpmtext{}s in homogeneous groups, such that the similarity is high within the individual groups and low between them. One clustering algorithm can produce different cluster sets for the same model set based on the parameters $P$. In our work, we focus on hierarchical clustering and consider distance threshold and linkage as possible parameters. In particular, we use linkage to determine how the distance between two clusters containing multiple models is calculated and the distance threshold to determine the maximum merging distance. For the traditional hierarchical clustering algorithm, we use the returned clusters in $\partition_{\lpm_L} = \textit{clust}_P(\lpm_L, \simmeasurevar_{\textit{name}})$ for an \lpmtext{} set $\lpm_L$ discovered on an event log $L$ and a similarity measure $\simmeasurevar_{\textit{name}}$ are pairwise disjoint, i.e., $\forall_{\lpm_i, \lpm_j \in \partition_{\lpm_L}} \lpm_i \cap \lpm_j = \emptyset$. We overload the notation $\partition_{\lpm_L}(\lpmvar)$ to denote the cluster in the cluster set $\partition_{\lpm_L}$ in which the \lpmtext{} $\lpmvar$ belongs. That is, it holds $\lpmvar \in \partition_{\lpm_L}(\lpmvar) \in \partition_{\lpm_L}$.

\subsection{Choosing Cluster Representatives (Step 3)}

In Step 3, we take a set of \lpmtext{}s $\lpm_L \subseteq \lpmuniverse$ discovered on an event log $L$ in Step 1, and a computed cluster set $\partition_{\lpm_L} = \textit{clust}(\lpm_L,\simmeasurevar_{\textit{name}})$ from Step 2. We return an \lpmtext{} set $\partition_{\lpm_L}^{\textit{repr}}$ in which we keep only one representative \lpmtext{} per cluster. In this work, we choose representative models either by taking the highest-ranked \lpmtext{} in each cluster considering some ranking function $\rank{L} \in \rankingfuncuniverse$ or the \lpmtext{} with the minimal mean distance to all other \lpmtext{}s in the cluster. In \Cref{def:repr}, we formally define \emph{representative projection} as a function that maps a set of \lpmtext{}s to one model.

\begin{definition}[Representative projection]
    \label{def:repr}
    A \emph{representative projection}
    $\textit{repr} \in \powerset{\lpmuniverse} \partmapsto 
    \lpmuniverse$ is a function that takes an \lpmtext{} set $\lpm_L$ and returns one representative \lpmtext{}. We use $\textit{repr}_{\textit{rank}}$ and $\textit{repr}_{\textit{dist}}$ to denote the representation projections based on the highest ranking and minimal mean distance respectively.
\end{definition}

Now, for the set of \lpmtext{}s $\lpm_L \subseteq \lpmuniverse$, we create the set $\partition_{\lpm_L}^{\textit{repr}_{x}} = \set{\textit{repr}_x(\lpm_i) \st \lpm_i = \partition_{\lpm_L}(\lpmvar) \land \lpmvar \in \lpm_L}$
and we call it the \emph{cluster representatives}, where $x \in \{\textit{rank}, \textit{dist}\}$.
This way, we significantly reduce the number of \lpmtext{}s from the original set $\lpm_L$, but still keep the essence of the entire set.

\section{Evaluation Results}\label{sec:evaluation_results}

The evaluation is performed on six \lpmtext{} sets, discovered on real event logs. In \Cref{tab:sets}, we give a summary of the \lpmtext{} sets and the corresponding event logs used in the experiments. 
For clustering, we use the agglomerative clustering algorithm of the \texttt{scikit-learn} package \cite{scikit-learn}. For all experiments, we use the complete linkage and iterate the distance thresholds between $0.1$ and $1.0$. For each combination of an \lpmtext{} set, similarity measure, and clustering algorithm parameter, we rerun the clustering $100$ times, resulting in $30000$ experiments. Whenever single values are shown, the most compact clustering according to the silhouette score was taken unless otherwise specified. All used cluster representatives we calculated using $\textit{repr}_\textit{dist}$.

\begin{table}[t]
    \centering
    \caption{Local process model sets used in the evaluation}
    \begin{tabular}{c|c|c}
        \hline
        \textbf{Event Log} & \textbf{LPM set} & \textbf{Number of models} \\ \hline
        BPI Challenge 2012 & $\lpm_{\textit{BPIC2012}}$ & $1096$  \\ \hline
        BPI Challenge 2012 - resource 10939 & $\lpm_{\textit{BPIC2012-res10939}}$ & $4496$  \\ \hline
        BPI Challenge 2017 &  $\lpm_{\textit{BPIC2017}}$ & $600$  \\ \hline
        Sepsis &  $\lpm_{\textit{Sepsis}}$ & $601$  \\ \hline
        Road Traffic Fine Management & $\lpm_{\textit{RTFM}}$ & $1694$ \\ \hline
        Hospital Billing &  $\lpm_{\textit{HB}}$ & $2051$  \\ \hline
    \end{tabular}
    \label{tab:sets}
\end{table}

Due to space limitations, in the remainder, we only show the results of some of the experiments. All analogous graphs (on other \lpmtext{} sets, measures, representative choosing strategies, or parameters), together with all resources needed to replicate the experiments, can be found on \url{https://github.com/VikiPeeva/CombiningLPMDandPMSM}.

\subsection{Case Study}
\begin{figure}
    \centering
    \includegraphics[width=0.3\linewidth]{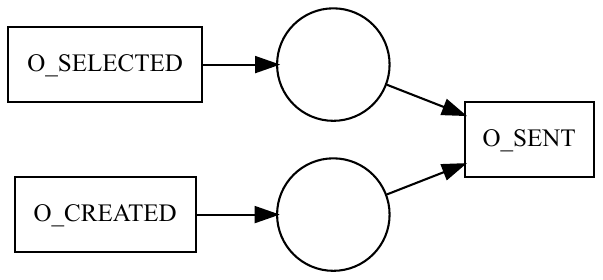}\hspace{1cm}
    \includegraphics[width=0.3\linewidth]{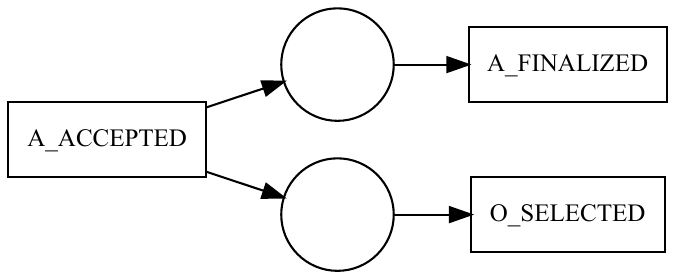} \\
    \vspace{0.5cm}
    \includegraphics[width=0.65\linewidth]{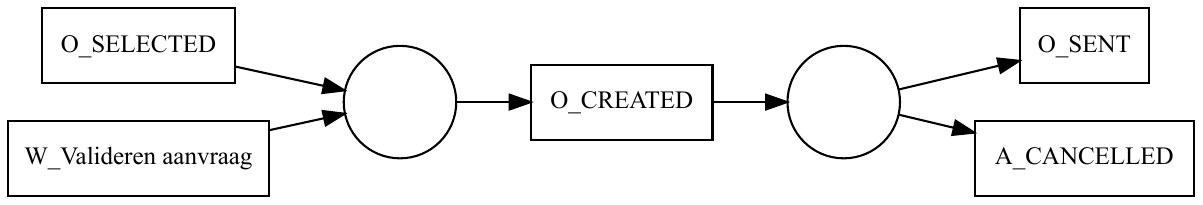}
    \caption{The three highest-ranked representative \lpmtext{}s for \constname{BPIC2012-res10939}.}
    \label{fig:highest_ranked_repr_models}
\end{figure}

We focus on the \constname{BPIC2012-res10939} event log in this part of the evaluation. In \cite{DBLP:journals/jides/TaxSHA16}, Tax et al. showed how we can use \lpmtext{}s to see different frequently appearing behavioral patterns that could not be seen on the start-to-end model because of too unstructured behavior. However, as shown in \Cref{fig:similar_models}, the highest-ranked models for both \cite{DBLP:journals/jides/TaxSHA16} and \cite{DBLP:conf/apn/PeevaMA22} focus on different behavioral variants of \constname{O\_SELECTED}, \constname{O\_CREATED}, and \constname{O\_SENT}. Such repetition appears for lower-ranked models as well. In \Cref{fig:highest_ranked_repr_models}, we show the three highest-ranked representative \lpmtext{}s after grouping the original set of \lpmtext{}s. It is clear that the behavioral span of these three models is significantly larger than the behavior described by the three highest-ranked original models discovered by both Peeva et al. and Tax et al (see \Cref{fig:similar_models}). If we map back the ranks of the three representative models to the original set, one would have to consider $1$, $17$, and $20$ higher ranked, but at the same time, more repetitive models before reaching them.

\subsection{Local Process Model Diversity Analysis}

In \Cref{sec:evaluation_setup}, we introduced how to reduce a set of \lpmtext{}s $\lpm$ to a subset of representative models $\partition_{\lpm}^{\textit{repr}}$ given a cluster set $\partition_{\lpm}$. In this part, we show how significant is the decrease from the original to the representative set for the most compact clusterings, and whether the set of $n$ highest-ranked \lpmtext{}s in the representative set is more diverse than the set of $n$ highest-ranked \lpmtext{}s in the  original set. 

\begin{figure}[h]
    \centering
    \includegraphics[width=\linewidth]{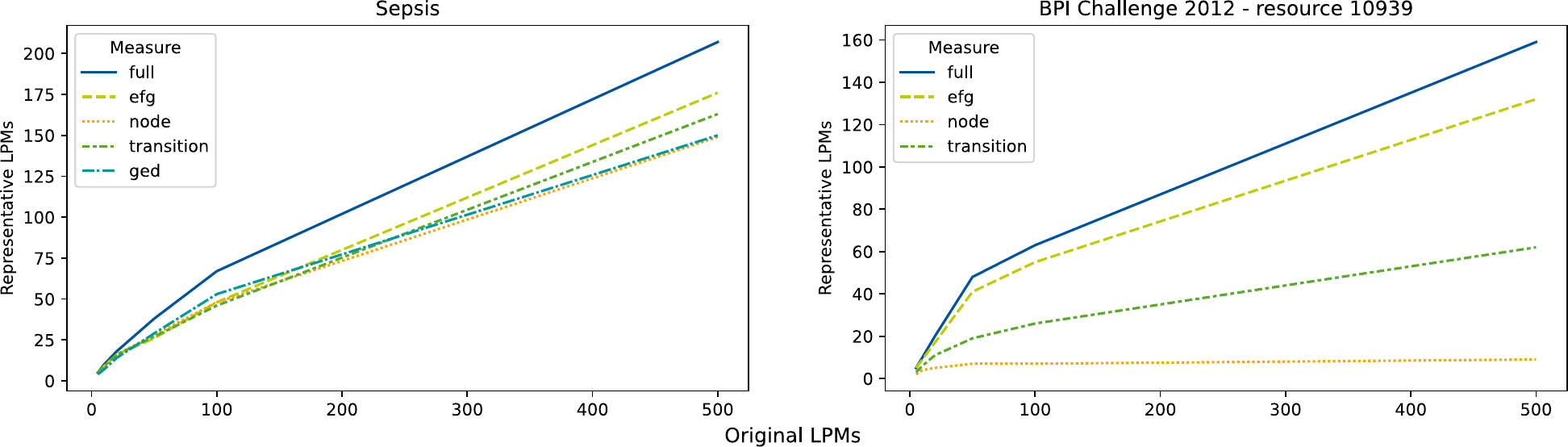}
    \caption{The number of models in the original set (x-axis) versus the number of representative models after clustering (y-axis).}
    \label{fig:eval_model_repetition}
\end{figure}

We start by considering the $5$, $10$, $20$, $50$, $100$, and $500$ highest-ranked \lpmtext{}s obtained on each of the event logs, that is, the sets $\lpm^{(5)}_L$, $\lpm^{(10)}_L$, $\lpm^{(20)}_L$, $\lpm^{(50)}_L$, $\lpm^{(100)}_L$, and $\lpm^{(500)}_L$, where $L$ represents the event logs in \Cref{tab:sets}.
In \Cref{fig:eval_model_repetition}, we illustrate the decrease in the number of \lpmtext{}s discovered on the \constname{BPIC2012-res10939} and \constname{Sepsis} event logs for the different similarity measures. The number $n$, denoting the $n$ highest-ranked models in the original set $\lpm^{(n)}_{L}$, is shown on the x-axis and the number of representative models in $\partition^{\textit{repr}_{\textit{dist}}}_{\lpm^{(n)}_{L}}$, is on the y-axis. It is clear that according to all similarity measures, the most compact clusterings of the \lpmtext{}s for the \constname{Sepsis} event log tend to have more clusters. Meaning, the original set already contains more diverse \lpmtext{}s, and correspondingly the decrease in the number of models is lower. On the contrary, when considering the \constname{BPIC2012-res10939} event log, it is noticeable that there is a mismatch between what different similarity measures consider compact grouping. The \textit{node} measure prefers a few clusters, while the \textit{full} measure favors much more clusters. Nevertheless, in both cases, the most conservative reduction still reduces the number of models by at least half. Additionally, it is worth mentioning that although these computations are done for the most compact clusterings in our experiments, in general, the number of clusters is something that can be controlled.

\begin{figure}[h]
    \centering
    \includegraphics[width=\linewidth]{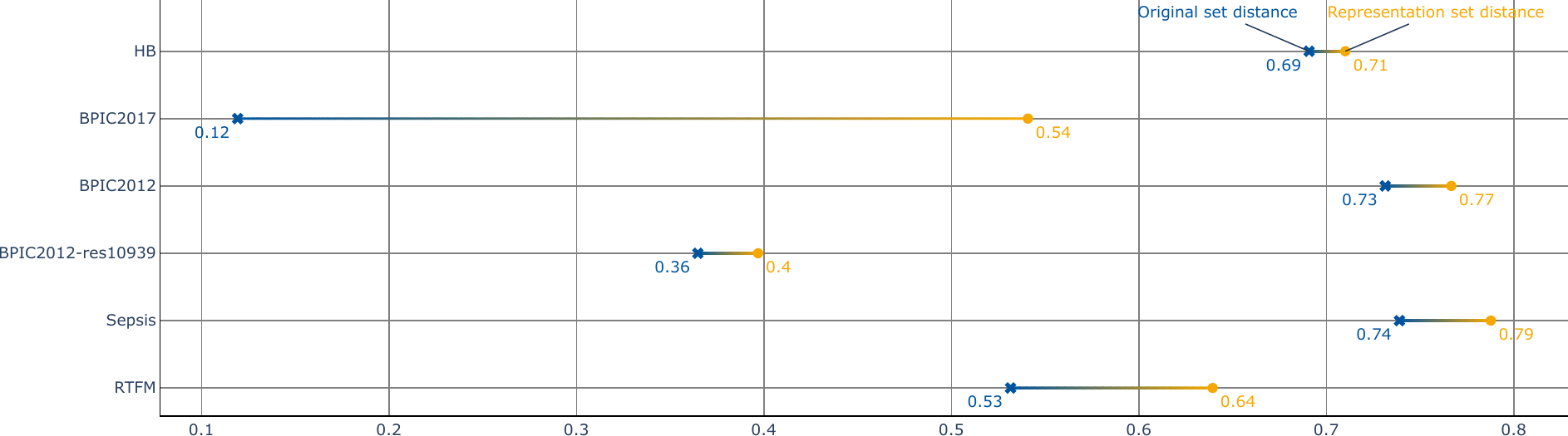}
    \caption{The mean distance between all pairs of \lpmtext{}s in the original set $\lpm^{(10)}_L$ versus the cluster representative set ${(\partition^{\textit{repr}_{\textit{dist}}}_{\lpm_{L}})}^{(10)}$ and the \emph{efg} measure.}
    \label{fig:diversity_orig_vs_repr}
\end{figure}

In the second part, we compare the mean distance between all pairs of the $n$ highest-ranked models in the original sets $\lpm^{(n)}_L$, versus the mean distance between all pairs of the $n$ highest-ranked models in the representative sets ${(\partition^{\textit{repr}_{\textit{dist}}}_{\lpm_{L}})}^{(n)}$ for $n=\{5, 10, 50, 100\}$. \Cref{fig:diversity_orig_vs_repr} shows the differences between $\lpm^{(10)}_L$ and ${(\partition^{\textit{repr}_{\textit{dist}}}_{\lpm_{L}})}^{(10)}$ for each of the event logs and the \textit{efg} measure. It is clear that in all cases the mean distance of the representative set is higher than the mean distance on the original set, meaning the set is more diverse. The highest increase can be noticed for the \constname{BPIC2017} event log and the smallest for the \constname{BPIC2012-res10939} event log. The distance increase happens for almost all $n$, measure and event log combinations, while for a few no significant change could be noticed.

\section{Conclusion}\label{sec:conclusion}
In this paper, we used process model similarity measures to group similar \lpmtext{}s together. We proposed a three-step approach consisting of \lpmtext{} discovery, clustering, and choosing \lpmtext{} cluster representatives. In the evaluation, we showed how grouping similar \lpmtext{}s together improves process understandability on a real-life case study and we showcased \lpmtext{} repetition decrease and diversity improvement on six real event logs.

There are numerous possibilites for future work. Currently, we experimented only on one \lpmtext{} discovery approach, hence, we can expand this work by considering \lpmtext{}s discovered with different algorithms. To further advance the method, one can also organize the \lpmtext{}s in each cluster set in hierarchies for more structured navigation between the models. Additionally, the framework could be extended with new similarity measures and different clustering algorithms. Finally, a natural extension would be to test whether \lpmtext{}s can be used to compare process model similarity measures in an unsupervised manner.
\end{sloppypar}

\section*{Acknowledgment}

\begin{minipage}[c]{\textwidth}
    We thank the Alexander von Humboldt (AvH) Stiftung for supporting our research.
    The authors gratefully acknowledge the financial support by the Federal Ministry of Education and Research (BMBF) for the joint project AIStudyBuddy (grant no. 16DHBKI016).
\end{minipage}

%
%
\bibliographystyle{splncs04}
\bibliography{bibliography}

\end{document}